\documentclass[twoside,twocolumn,10pt]{article}
\usepackage{wscg}           % includes a number of other packages (e.g., myalgorithm)
\RequirePackage{ifpdf}
\ifpdf
 \RequirePackage[pdftex]{graphicx}
 \RequirePackage[pdftex]{color}
\else
 \RequirePackage[dvips,draft]{graphicx}
 \RequirePackage[dvips]{color}
\fi

\usepackage{nopageno}       % no page numbers at all; uncomment for final version

\usepackage{comment}

\title{Radial Line Fourier Descriptor for Historical \\Handwritten Text Representation}
\author{
\parbox{0.55\textwidth}{\centering
Anders Hast and Ekta Vats\\[1mm]
Department of Information Technology\\
Uppsala University, SE-751 05 Uppsala, Sweden\\[1mm]
anders.hast@it.uu.se; ekta.vats@it.uu.se
}}

\usepackage{url}
\makeatletter
\def\Uslash{\mathbin{\mathchar`\/}\@ifnextchar{/}{\kern-.15em}{}}
\g@addto@macro\UrlSpecials{\do \/ {\Uslash}}
\def\Ucolon{\mathbin{\mathchar`:}\@ifnextchar{/}{\kern-.1em}{}}
\g@addto@macro\UrlSpecials{\do : {\Ucolon}}
\makeatother

\begin{document}

\twocolumn[{\csname @twocolumnfalse\endcsname

\maketitle  % full width title

\begin{abstract}
\noindent
Automatic recognition of historical handwritten manuscripts is a daunting task due to paper degradation over time. Recognition-free retrieval or word spotting is popularly used for information retrieval and digitization of the historical handwritten documents. However, the performance of word spotting algorithms depends heavily on feature detection and representation methods. Although there exist popular feature descriptors such as Scale Invariant Feature Transform (SIFT) and Speeded Up Robust Features (SURF), the invariant properties of these descriptors amplify the noise in the degraded document images, rendering them more sensitive to noise and complex characteristics of historical manuscripts. Therefore, an efficient and relaxed feature descriptor is required as handwritten words across different documents are indeed similar, but not identical. This paper introduces a Radial Line Fourier (RLF) descriptor for handwritten word representation, with a short feature vector of 32 dimensions. A segmentation-free and training-free handwritten word spotting method is studied herein that relies on the proposed RLF descriptor, takes into account different keypoint representations and uses a simple preconditioner-based feature matching algorithm. The effectiveness of the RLF descriptor for segmentation-free handwritten word spotting is empirically evaluated on well-known historical handwritten datasets using standard evaluation measures.

\end{abstract}

\subsection*{Keywords}
Radial Line Fourier descriptor, word spotting, feature matching
\vspace*{1.0\baselineskip}
}]

%%%%%%%%%%%%%%%%%%%%%%%%%%%%%%%%%%%%%%%%%%%%%%%%%%%%%%%%%%%%%%%%%%%%%%%%%%%%%

\section{Introduction}
%\copyrightspace
Automatic recognition of poorly degraded handwritten text is challenging due to complex layouts and paper degradations over time. Typically, an old manuscript suffers from degradations such as paper stains, faded ink and ink bleed-through. There is variability in writing style, and the presence of text and symbols written in an unknown language. This hampers the document readability, and renders the task of searching a word in a set of non-indexed documents i.e. word spotting, to be more difficult. 

In literature \cite{gio17}, word spotting approaches can either be segmentation-based where the search space consists of a set of segmented word images, or segmentation-free with the complete document image in the search space. This paper focuses on segmentation-free word spotting, which is typically preferred over segmentation-based methods when dealing with heavily degraded document images \cite{zag17}. However, the performance of word spotting algorithms significantly depends on the appropriate selection of feature detection and representation methods \cite{gio17}. In general, feature descriptors represent a region with distinct feature in a document image, coded into a numerical feature vector, which is subsequently compared with the feature vector of a reference image to perform matching. 

Efforts have been made in the recent past towards research on feature detection and representation methods. Some popular methods include Scale Invariant Feature Transform (SIFT) \cite{low04}, Speeded Up Robust Features (SURF) \cite{bay08} and Histograms of oriented Gradients (HoG) \cite{dal05}. SIFT and HoG contributed significantly towards the progress of several visual recognition systems in the last decade \cite{gir14}. However, these local descriptors were mainly designed for the representation of natural scene images, that possess structurally different characteristics from the document images. For example, the detection of the most important edges using pyramid scaling in SIFT creates local interest points between the text lines \cite{zag17}. The invariant properties of these descriptors amplify the noise in the degraded document images, rendering them more sensitive to noise and complex characteristics of historical manuscripts \cite{zag17}. The work by \cite{ley09} analyzed that the rotation-invariant features are more sensitive to noise in a document image, and perform poorly as compared to rotation-dependent features.

Since the existing descriptors are found to be unsuitable for representing handwritten text with high levels of degradations \cite{zag17,ley09}, it is important to design a descriptor to address this issue. This paper introduces a Radial Line Fourier (RLF) descriptor which is tailor-made for word spotting applications with fast feature representation and robustness to degradations. RLF is a fast and short-length feature vector of 32 dimensions, based on log-polar sampling followed by computing a few elements of the Discrete Fourier Transform (DFT) along each radial line. It does not require any orientation information from the feature detectors, and simple feature detectors can be used without compromising the descriptor and word spotting performance.

This paper is organized as follows. Section \ref{sec:Feat} reviews the state-of-art methods used in word spotting pipeline, with main focus on interest point detection and feature representation methods. Section \ref{sec:method} presents the proposed method based on the RLF descriptor for segmentation-free handwritten word spotting. Section \ref{sec:exp} demonstrates the efficacy of the proposed method on well-known historical datasets using standard evaluation measures. Section \ref{sec:con} concludes the paper.

\section{Related Work}
\label{sec:Feat}

Appropriate selection of interest points (keypoints) and feature descriptors is indispensable for the performance of a word spotting system. This section discusses some popular interest point detection and feature representation methods with reference to word spotting systems. It is important to note that the segmentation-free word spotting framework presented herein is training-free, therefore training-based methods such as deep learning are not considered.

\subsection{Interest Point Detection}

Feature detection, or interest point detection refers to finding keypoints in an image that contain crucial information. There exist several interest point detectors in literature. For example, the Harris corner detector \cite{har88} is popularly used for corner points detection. It computes a combination of eigenvalues of the structure tensor such that the corners are located in an image. Shi-Tomasi corner detector \cite{shi94} is a modified version of Harris detector. The minimum of two eigenvalues is computed and a point is considered as a corner point if this minimum value exceeds a certain threshold. The Maximally Stable Extremal Regions (MSER) \cite{mat02} detector detects keypoints such that all pixels inside the extremal region are either darker or brighter than all the outer boundary pixels.

Typically, interest point based feature matching is performed by using a single interest point detector type. SIFT and SURF are the most popular detectors that capture the blob type of features in the image. SIFT uses the Difference of Gaussians (DoG) that computes the difference between Gaussian blurred images using different values of $\sigma$, where $\sigma$ defines the Gaussian blur from a continuous point of view. SURF computes the Determinant of the Hessian (DoH) matrix, that defines the product of the eigenvalues. In principle, any combination of different keypoint detectors can be selected depending upon the application. This work uses a combination of four types of keypoint detectors for handwritten text representation, that consists of corner detectors, dark and bright blobs, saddle points, and the edges of text strokes.

\subsection{Feature Representation}

After a set of interest points has been detected, a suitable representation of their values has to be defined to perform word matching. In general, a feature descriptor is constructed from the pixels in the local neighborhood of each interest point. Fixed length feature descriptors are most commonly used that generate a fixed length feature vector, which can be easily compared using standard distance metrics (e.g. the Euclidean distance). Sometimes, fixed length feature vectors are computed directly from the extracted features without the need of a learning step \cite{gio17}.

Gradient-based feature descriptors tend to be superior, and include SIFT \cite{low04}, HoG \cite{dal05} and SURF \cite{bay08} descriptors. The 128-dimensional SIFT descriptor is formed from histograms of local gradients. SIFT is both scale and rotation invariant, and includes an intricate underlying framework to ensure this. Similarly, HoG computes a histogram of gradient orientations in a certain local region. An important difference between SIFT and HoG is that HoG normalizes the histograms in overlapping blocks, and creates a redundant expression. SURF descriptor is generally faster than SIFT, and is created by concatenating Haar wavelet responses in sub-regions of an oriented square window. SIFT and SURF are invariant to both scale and rotation changes. There are several variants of these descriptors that have been employed for word spotting \cite{rod08,gio17}.

Many feature descriptors use local image content in square areas around each interest point to form a feature vector \cite{has16}. Both scale and rotation invariance can be obtained in different ways \cite{gau11}. The Fourier transform has been used to compute descriptors that is illumination and rotation invariant, and scale-invariant to a certain extent \cite{car02,car03}. In order to overcome dimensionality issues that may arise in a high-dimensional space, binary descriptors are introduced that are faster, but less precise, for example the Binary Robust Invariant Scalable Keypoints (BRISK) descriptor \cite{leu11} and Fast Retina Keypoint (FREAK) descriptor \cite{ala12}. 

However, these descriptors with strict invariance properties are not suitable for handwritten document representation. This is mainly because the invariance property renders them more sensitive to noise in a degraded document, as has been carefully studied in \cite{zag17,ley09}. 

A method for searching handwritten Arabic documents based on a set of binary shape features is presented in \cite{sri05}, where a correlation distance based matching technique has been employed. However, it was argued by \cite{gau11} that the features that are dependent on word shape characteristics are not effective in dealing with multi-writer document collections. Instead, the texture information in a spatial context is considered more reliable than the shape information, as suggested in \cite{gau11,lla12}.

In \cite{ley07}, the image zones representing the most informative parts in a document image are detected based on the gradient orientation computed by taking convolution of the image with the first and second derivatives of the Gaussian kernel. However, this method was found to be inefficient for short words with less than four characters, and therefore an improved version was proposed in \cite{ley09}. The feature matching algorithm in \cite{ley09} was found to be very sensitive to variations in handwriting and font sizes, and the overall matching process was too slow for processing large datasets. An interesting block-based document image descriptor was presented by \cite{gat09} where the query image was scaled and rotated to produce different word instances, and for each instance, a different set of feature vectors was computed. However, several versions of queries generated significant amount of noise in the final merging state, rendering the method inefficient for handling large writing style and font variations.

Inspired by Bag-of-Visual Words (BoVW) model, a patch-based framework that uses SIFT for local feature representation was presented in \cite{rus11}. The codebook generation step of BoVW model is expensive, and this method is also found to be unsuitable for handling query font size and handwriting variations \cite{zag17}. The performance of popular word descriptors in a BoVW context was evaluated in \cite{lla12}, and it was suggested that the statistical BoVW approach generates the best result, but with significant increase in overhead in terms of memory requirements to store the descriptors.

The winning algorithm, \cite{kov14}, for segmentation-free track of ICFHR 2014 Handwritten Keyword Spotting Competition \cite{pra14}, employed HoG and Local Binary Patterns (LBP) descriptors, and the word retrieval is performed using the nearest neighbour search, followed by a simple oppression of extra overlapping candidates. The work by \cite{zag17} outperformed the winning algorithms from ICFHR 2014 Handwritten Keyword Spotting Competition \cite{pra14}, and ICDAR2015 Handwritten Keyword Spotting Competition \cite{pui15}. They proposed a new approach towards handwritten word spotting, where the spatial information representing the current location of a feature point is taken into account, and is based on the texture information. However, it is unclear how well this method performs in challenging cases where a a word shares several letters with other different words. The RLF descriptor based method proposed herewith handles this issue by dividing a word into several parts (depending upon the size of the word) to eliminate false-positives, and perform reliable keypoint-based feature matching.

\begin{figure*}[htb]
\centering
\includegraphics[scale=.45]{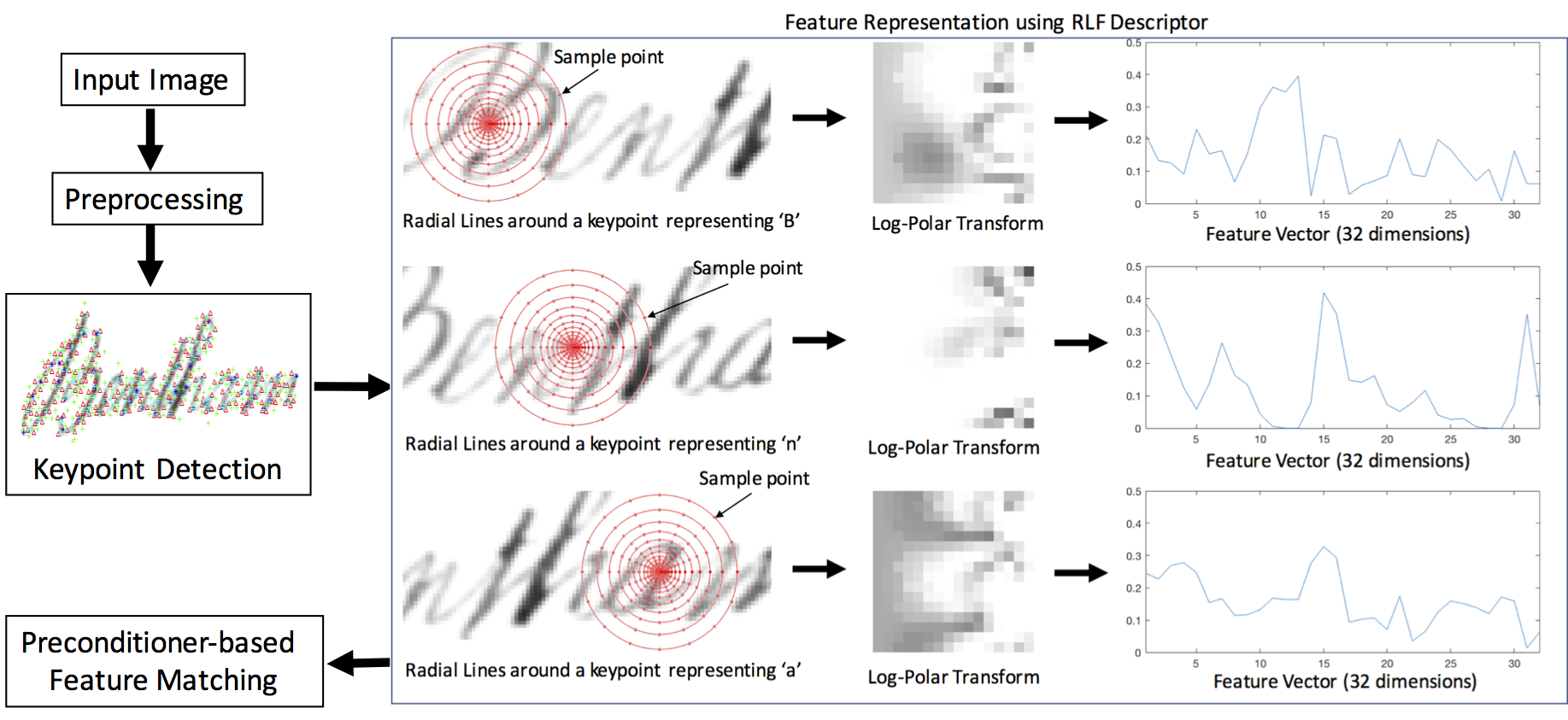}
\caption{Radial Line Fourier (RLF) descriptor for feature representation in a word spotting framework. Each keypoint detected is represented using log-polar sampling scheme with 16 sampling points per ring. Each radial line, originating in the center and traversing each ring, is used to obtain a square (16 x 16) transformed image representation. In the next step, DFT is applied along each row (corresponding to radial lines) to compute the amplitude of a few elements for each row that constitute the feature vector. Finally, the feature vector generated is presented where x-axis denotes the feature vector length (i.e. 32), and y-axis denotes the amplitude of DFT.}
\label{fig_rlf}
\end{figure*}

The performance of different features for word spotting applications was evaluated using Dynamic Time Warping (DTW) \cite{rod08} and Hidden Markov Models (HMMs) \cite{rod09}. It was found that the local gradient histogram features outperform other geometrical or profile-based features. These methods generally match features from evenly distributed locations over normalized words where no nearest neighbor search is necessary. This is because each point in a word has its corresponding point in some other word located in the very same position. Recently, a method based on feature matching of keypoints derived from the words was proposed \cite{has16}, which requires a nearest neighbor search. In this case, a relaxed descriptor is required that is not over-precise, since the handwritten words are not normalized. This is due to complex characteristic of handwritten words, unlike simple Optical character recognition (OCR) text. Handwritten words across different documents are similar, but not identical due to variability in writing styles.

In an endeavor to address the issues discussed above, this work proposes the RLF descriptor, which is tailor-made for handwritten words representation. The main highlights of this work are as follows: (a) a segmentation-free and training word spotting approach is studied; (b) the proposed method uses a combination of different keypoint detectors to capture different characteristics in a handwritten document, which consists of both lines, corners and blobs; (c) the RLF descriptor is designed, which is a fast and short-length feature vector of 32 dimensions with several advantages; (d) a simple preconditioner-based feature matching algorithm is presented. Advantages of RLF descriptor include faster word spotting (due to short length of feature vector), robustness to degradations, flexibility to be employed with existing feature detectors, efficient memory utilization, and no increase in overhead for feature orientation estimation. The proposed methodology is discussed as follows.

\section{Methodology}
\label{sec:method}

The pipeline of the word spotting framework is as follows. For an input document image, preprocessing is performed to remove background noise using two band-pass filtering approach \cite{vat17}. This is followed by keypoints detection, feature representation using RLF descriptor, and preconditioner-based feature matching. The framework of the proposed approach is pictorially described in Figure \ref{fig_rlf}.

\subsection{Preprocessing}

Preprocessing is the initial step of the word spotting algorithm where the background noise is removed using a simple two band-pass filtering approach, as proposed in \cite{vat17}. A high frequency band-pass filter is used to separate the fine detailed text from the background, and a low frequency band-pass filter is used for masking and noise removal. The background removal is performed in such a way that the gray-level information crucial for the feature extraction is not affected. This allows the keypoint detector and the RLF descriptor to be more informative.

\subsection{Keypoint Detection}

\begin{figure}[htb]
    \centering
    \includegraphics[scale=.34]{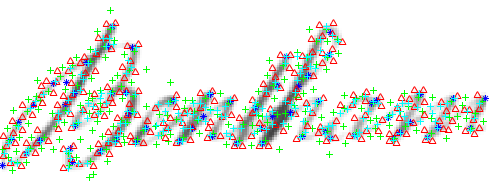}
    \caption{An example of a query word $Bentham$ depicting four different types of keypoints. Blue * is a corner detector, green + finds the dark and bright blobs, red $\Delta$ finds the saddle points, and cyan + finds the edges of the text strokes.}
    \label{Fig_kpExample}
\end{figure}

To begin with, keypoints are detected for the document image and the query word. A combination of four different types of keypoint detectors is used to capture a variety of features that represent a handwritten document, and consists of lines, corners and blobs. Figure \ref{Fig_kpExample} presents the keypoint detectors used herein using an example image of a smoothed query word, $Bentham$. $Blue$ * represents the Harris corner detector \cite{har88}, $green$ + represents the result of using the square of the Determinant of Hessian (DoH), which captures both dark and bright blobs, $red$ $\Delta$ represents negative of DoH (-DoH) and finds the saddle points, and $cyan$ + represents the result of an edge detector ($Assymetric^2$) \cite{has14b}.

\subsection{Radial Line Fourier Descriptor}

Radial Line Fourier (RLF) descriptor is a short-length feature vector of 32 dimensions, presented in this work for representation of handwritten words. RLF is inspired from a variant of Scale Invariant Descriptor (SID) \cite{kok08}, known as SID-Rot \cite{tru13}, and the idea is to perform log-polar sampling in a circular neighborhood around each keypoint. SID is a scale and rotation invariant descriptor, whereas SID-Rot is scale-invariant but rotation-sensitive descriptor. Typically, the Fourier transform can be applied over scales only to obtain a scale-invariant and rotation-dependent descriptor, or a rotation-invariant and scale-sensitive descriptor. The Fourier transformation over scales render the SID-Rot to be rotation-sensitive, and the scale invariance is achieved by sampling over a large radius with a descriptor length of $3360$. This method works well in representing natural scene images with scale changes and no rotations. However, strict invariance properties amplify noise in degraded document images \cite{zag17}, and may lead to loss of useful information. Therefore, a relaxed feature descriptor, such as RLF, is required.

RLF descriptor computes a feature vector representation of an image feature, and is based on log-polar sampling followed by computing a few elements of the DFT along each radial line. It characterizes an image region as a whole using a single feature vector of fixed size, and no learning step is involved. Figure \ref{fig_rlf} presents the general framework of the RLF descriptor for feature representation in a word spotting pipeline, and discussed in detail as follows.

After the keypoints representing a document image have been detected, log-polar sampling is performed at each keypoint, where each radial line (going from the center, traversing each ring around the center along a line) is transformed into a square representation, as highlighted in Figure \ref{fig_rlf}. The log-polar transform resampling resolution is set to 16 sample points per ring to obtain a square (16 x 16) transformed image. When sampling is done in a log-polar fashion, certain interpolation is required as the pixel coordinates are seldom in the center. One could for instance use a bilinear interpolation to achieve higher accuracy. In this work, sub-pixel sampling is computed using the Gaussian interpolation in a 3x3 neighborhood. In the next step, DFT is used to compute the amplitude of a few elements that constitute the feature vector.

The Fast Fourier Transform (FFT) performed efficiently in \cite{has14a} for creating descriptors that are relaxed. However, it was found to be impractical for high level applications with large amount of data \cite{has16}. This is because the FFT is rather slow in computations, such as computing the distance measures (i.e. phase correlation). In general, the FFT requires $O\left(N ~ log(N)\right)$ computations for a discrete series $f(n)$ with $N$ elements. Therefore, this work improves and simplifies the computations needed to generate a faster feature representation, still benefiting from the advantages of the Fourier transform. We propose to use just a few elements from DFT of the sampled elements $f(n)$ along the radial line, and the computation required (using Euler's formula) is 
\begin{equation}
\small
\mathcal{F}[f(n)](k)=\sum_{n=0}^{N-1}{f(n) \cos(2 \pi n k /N)- i (f(n) \sin(2 \pi n k /N))}.
\end{equation}
The value of $k$ determines the frequency used to compute the Fourier element, where $k \in {0, 2, 4..}$. Typically, noise in a document image has higher frequency as compared to the main text in the document image, therefore the second ($k=2$) and third ($k=4$) elements of the Fourier transform are selected to form the feature descriptor. DFT requires only $O\left(N\right)$ computations per element. Note that the Discrete Cosine (DC) component is obtained for $k=0$ and is less informative. The trigonometric functions in the DFT do not have to be computed for each step, and the computation requires simple mathematical operations using the Chebyshev recurrence relation, same as the original Fourier Transform.

The RLF descriptor is thus constructed by computing the amplitude of a few elements of DFT:
\begin{equation}
\left| \mathcal{F}[f(n)](k) \right|= \sqrt{\Re{\left(\mathcal{F}[f(n)](k)\right)}^2 + \Im{\left(\mathcal{F}[f(n)](k)\right)}^2}.
\end{equation}
The descriptor computation using only $k=2$ suffices well for handwritten word representation under the test settings, and most importantly the descriptor is very short (length 32) with fast feature representation. However, experimentally it was found that by adding a second element for $k=4$, the quality of the subsequent matching improved, even though the feature vector thus generated is twice as long. The advantage is that it makes it possible to sample in a smaller neighborhood, while still getting the same number of corresponding matches, with better accuracy. Nevertheless, adding a third element for $k=6$ did not improve the accuracy significantly, and is found to be not worth the extra computational effort. This work uses RLF descriptor with length 32 for experimental analysis, taking into account the trade-off between computational cost and accuracy.
% Comment: Can the quality of matching be problem dependent? And in turn, the best value of k be problem dependent? You can discuss/comment on this here?

The RLF feature vector thus generated is presented in Figure \ref{fig_rlf}, where x-axis denotes the feature vector length (i.e. 32 dimensions), and y-axis denote the frequency amplitudes of DFT. The advantages of the RLF descriptor are many-fold. The RLF descriptor computes a fast and short-length feature vector, to be able to perform quick feature matching in the nearest neighbor search. The RLF descriptor emphasizes on the pixels closer to the feature center, making it less sensitive to erroneous feature size estimation. It is resistant to high frequency changes, such as due to residuals from neighboring words, as it is based on the low frequency content in the local neighborhood. Nevertheless, it is insensitive to small differences in form and shape, as long as they are almost same, i.e. the low frequencies are sufficiently similar.

\subsection{Feature Matching}

\begin{figure}[!t]
\centering
\includegraphics[scale=.15]{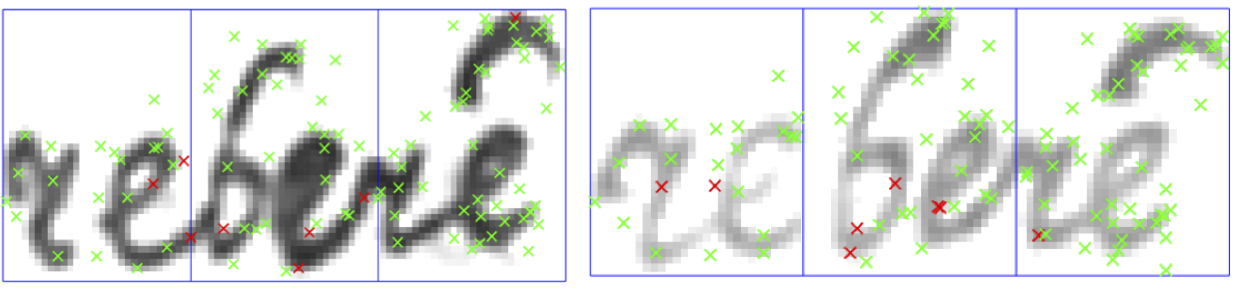}
\caption{Matched points represented using the RLF descriptor for a sample word {\it{reber\'e}}. The matching keypoints (inliers) are in green, and the matches discarded (outliers) by the preconditioner are in red.}
\label{Fig_kpExample2}
\end{figure}

A segmentation-free and training-free word spotting method based on the proposed RLF descriptor is studied herein. In general, no prior information is available about the potential word in the document that is to be matched with the query word. By using the RLF descriptor, the word matching problem is reduced to a much faster search problem. In this work, a simple preconditioner-based feature matching algorithm is employed.

To begin with, words are partitioned into several parts in order to avoid confusion between similar words and reduce false positives. This is to overcome the drawback of keypoint-based matching techniques \cite{zag17}, where parts of the retrieved words may be very similar to some part of the query word, or where a word shares several letters with other different words, hence generate false positives. In the experiments, words are divided into several parts depending upon the length of the query word. For example, in Figure \ref{Fig_kpExample2}, a sample query word {\it{reber\'e}} and its corresponding retrieved word are divided into three parts, and the preconditioner-based matching is performed in the respective three different parts of both the words.

After the partitioning step, a nearest neighbor part-based search is performed in an optimal sliding window within the subgroups of the detected keypoints. The keypoint matching algorithm computes the extent of the the matching points in a word, and therefore is able to capture words that are partially outside the sliding window. Consequently, the matched points are removed from the set of points when a word is found, to avoid finding the same word again.

The resultant correspondences between the query word and the retrieved word in the sliding window obtained after a simple keypoint matching consists of many outliers and needs further refinement. A common approach is to use Random sample consensus (RANSAC) \cite{fis87} to learn transformations between the words. However, it is important to have a relaxed transformation instead, because the same word at different locations in a document can differ with small variations in font sizes, or even larger variations in a multi-writer scenario. Therefore, a deterministic preconditioner (inspired from \cite{has12a}) is used in this work that eliminates the need to use RANSAC and helps in removing the false matches. In \cite{has16}, preconditioner had been used along with Putative Match Analysis (PUMA) \cite{has12b}, which is found to be computationally expensive and increases overhead in computing false positives. To keep the matching algorithm simple yet effective, this work uses a matching algorithm that is solely based on the preconditioner.

The preconditioner creates a cluster of corresponding matches in a two-dimensional space as positional vectors. This means that the correspondences between the query word and the retrieved word in the sliding window with same length and direction are potential inliers, that forms a two-dimensional cluster. However, the clusters are expected to be slightly scattered due to complex characteristics of words (e.g. words can differ in font and style), therefore the threshold must be relaxed or loosely set. The preconditioner finds the inliers efficiently and removes the outliers with fast computation speed. Figure \ref{Fig_kpExample2} represents the matched points obtained from the proposed method, where the matching keypoints or inliers are highlighted in green and the outliers discarded by the preconditioner are in red. The preconditioner-based matching efficiently captures complex variations in handwriting by estimating the core text dimensions on-the-fly. The effectiveness of the proposed method has been experimentally demonstrated in the next section.

\section{Experimental Results}
\label{sec:exp}

This section describes the datasets used in the experiments, and empirically evaluates the proposed method.

\subsection{Datasets}
For experimental analysis, the Barcelona Historical Handwritten Marriages dataset, and the Bentham dataset in two variants are taken into account. The former is heavily degraded, posing challenges for the word spotter, and the latter in both variants demonstrate multi-writer handwriting variations to a certain extent, along with document degradations. The datasets are discussed as follows:
\begin{itemize}
 \item {\it{Barcelona Historical Handwritten Marriages Dataset (BH2M)}}: It consists of historical handwritten marriage records stored in the archives of Barcelona cathedral, written between 1617 and 1619 by a single writer in old Catalan. The reader is referred to \cite{fer14} for a deeper understanding of the dataset.
 \item {\it{Bentham Dataset}}: It consists of handwritten document pages from the Bentham collection, which have been prepared in the $tranScriptorium$ project. The Bentham collection consists of manuscripts on law and moral philosophy handwritten by Jeremy Bentham (1748-1832) over a period of 60 years, and some handwritten documents from his secretarial staff. This dataset in first variant was used in ICFHR 2014 Handwritten Keyword Spotting Competition \cite{pra14}, and the second variant in ICDAR 2015 Handwritten Keyword Spotting Competition \cite{pui15}. For the experiments, all pages from both variants of Bentham dataset used in the competitions are employed, which have been written by different authors in different styles, font-sizes, and contains crossed-out words.
\end{itemize}

\subsection{Results}

The performance of the proposed method is empirically evaluated against the winning algorithms of ICFHR 2014 Handwritten Keyword Spotting Competition \cite{pra14}, and ICDAR2015 Handwritten Keyword Spotting Competition \cite{pui15}, along with the other state-of-the-art methods such as \cite{zag17,ley09}. The evaluation measure used is the classic mean Average Precision (mAP) metric popularly used in document word spotting. In general, the retrieved regions of all the document pages are combined and re-ranked according to the score obtained. If a region overlaps more than 50\% of the area of the ground truth corpora, it is classified as a positive region. The Precision and Recall values are first computed, and since a single value is preferable for comparison across different methods, the mAP of each method is calculated as the final result. A higher value of mAP is more desirable.

\begin{table}[htb]
 \centering 
  \begin{tabular}{|l|l|}
	\hline 
	\textbf{Method} & \textbf{mAP} \\
    \hline 
	\cite{alm12b} & 0.513 \\
	\hline 
    \cite{zag17} & 0.530 \\
	\hline 
    \textbf{Proposed method} &  \textbf{0.783}\\
	\hline 
  \end{tabular}
 \caption{Experimental results for BH2M Dataset.}
 \label{tab:barcelona}
\end{table}

\begin{table}[htb]
 \centering 
  \begin{tabular}{|l|l|}
	\hline 
	\textbf{Method} & \textbf{mAP} \\
    \hline 
    \cite{ley09} & 0.221 \\
	\hline 
	\cite{how13} & 0.409 \\
     \hline 
     \cite{kov14} & 0.423 \\
     \hline 
    \cite{zag17} & 0.517 \\
	\hline 
    \textbf{Proposed method} &  \textbf{0.490} \\
	\hline 
  \end{tabular}
  \caption{Experimental results for Bentham Dataset used in ICFHR 2014 competition.}
 \label{tab:bentham2014}
\end{table}

\begin{table}[htb]
 \centering 
  \begin{tabular}{|l|l|}
	\hline 
	\textbf{Method} & \textbf{mAP} \\
    \hline 
	PRG, TU Dortmund & 0.293 \\
     \hline 
	CVC, Spain &  0.116\\
	\hline 
    \cite{zag17} & 0.326 \\
	\hline 
    \textbf{Proposed method} &  \textbf{0.786} \\
	\hline 
  \end{tabular}
  \caption{Experimental results for Bentham Dataset used in ICDAR 2015 competition.}
 \label{tab:bentham2015}
\end{table}

Tables \ref{tab:barcelona}-\ref{tab:bentham2015} present the segmentation-free handwritten word spotting results for various methods. In Table \ref{tab:barcelona}, the performance of the proposed method is evaluated on the BH2M dataset against the methods proposed in \cite{alm12b} and \cite{zag17}. The method proposed in \cite{alm12b} is based on exemplar-SVM framework for word spotting, and the method presented in \cite{zag17} is based on Document-oriented Local Features (DoLF). It is observed from Table \ref{tab:barcelona} that the proposed method achieves higher mAP as compared to \cite{alm12b} and \cite{zag17}. This is mainly because the performance of \cite{alm12b} and \cite{zag17} is found to be weaker for challenging cases where a a word shares several letters with other different words. Typically, a higher mAP is achieved when search is performed on a long query word (e.g. $habitant$), as there is less possibility of finding the query word as part of other similar word. However, in an ideal scenario it is highly possible for a query word to share several characters with other words, even with a longer word. A simple example of a query word from the BH2M dataset is $donsella$, where some characters are common with query words $fill$ and $filla$. A much challenging case observed is the sequence of overlapping characters in the query words $fill$ and $filla$, where $fill$ is retrieved while searching for $filla$. The proposed method handles this effectively by dividing a word into several parts depending upon the length of the word, and then perform part-based keypoint matching. This simple approach reduces the false-positives by a significant margin, as is evident from the results in Table \ref{tab:barcelona}.
% (e.g. while searching for $filla$, $fill$ will not be retrieved)

Table \ref{tab:bentham2014} presents the results obtained using different methods on the Bentham dataset from ICFHR 2014 Handwritten Keyword Spotting Competition \cite{pra14}. The performance of the proposed method is empirically evaluated against the state-of-the-art methods such as \cite{ley09}, \cite{how13}, \cite{kov14} (i.e. winner of ICFHR 2014 competition), and \cite{zag17}. It is observed from Table \ref{tab:bentham2014} that the proposed method achieves higher mAP as compared to \cite{ley09}, \cite{how13} and \cite{kov14}, and performs comparable against \cite{zag17} for all test images under the experimental settings. This is mainly because the relaxed nature of RLF allows it to capture more details in a degraded document image as compared to descriptors with stricter invariance properties that render them more sensitive to noise. This is important as the same query word at different locations in a document can differ with small variations in font sizes, or even larger variations in a multi-writer scenario. However, even though the proposed approach is observed to perform significantly in comparison with other methods discussed in Table \ref{tab:bentham2014}, a mAP of 0.490 suggests further investigation. It is observed that the document images in the Bentham dataset from ICFHR 2014 competition consists of handwritten text from two or more authors, where the core text size in a document page differs across different locations in the same document page. This pose challenges for the algorithm in estimating the average core text size for each document page, as the normalization of text size might result in loss of information. The authors aim at investigating this issue further and working towards the improvement of the proposed algorithm as future work.

Table \ref{tab:bentham2015} evaluates the performance of the proposed method on the second variant of Bentham dataset introduced in the ICDAR 2015 Handwritten Keyword Spotting Competition \cite{pui15}. Unlike the first variant of the Bentham dataset discussed above, this dataset does not significantly suffer from the problem of highly variable core text size across a document page. This is evident from the higher mAP value achieved in Table \ref{tab:bentham2015}. It is observed that the proposed method achieves higher accuracy in comparison with the winner algorithms from the competition, as well as a recent method \cite{zag17}. The RLF descriptor with relaxed feature description takes into account the handwriting variations to a considerable extent, and the standard core text size is estimated for each document page without significant errors. %The reader is referred to the Matlab implementation of the proposed method with respect to the Bentham dataset used in ICDAR 2015 competition at \url{http://bit.ly/2tNm4Gr} to further validate the results in Table \ref{tab:bentham2015}.

\begin{table}[htb]
 \centering 
  \begin{tabular}{|l|l|}
	\hline 
	\textbf{Method} & \textbf{mAP} \\
    \hline 
	SIFT \cite{low04} & 0.115 \\
     \hline 
	SURF \cite{bay08} &  0.106\\
    \hline 
	BRISK \cite{leu11} & 0.035 \\
    \hline 
	ORB \cite{rub11} & 0.098 \\
    \hline 
	KAZE \cite{alc12} & 0.283 \\
	\hline 
    DoLF \cite{zag17} & 0.517 \\
	\hline 
    \textbf{Proposed RLF} &  \textbf{0.490} \\
	\hline 
  \end{tabular}
   \caption{Performance evaluation of feature representation methods on Bentham Dataset used in ICFHR 2014 competition.}
 \label{tab:feat1}
\end{table}

\begin{table}[htb]
 \centering 
  \begin{tabular}{|l|l|}
	\hline 
	\textbf{Method} & \textbf{mAP} \\
    \hline 
	HoG \cite{alm12a} & 0.584  \\
     \hline 
	Loci \cite{fer11} & 0.419 \\
    \hline 
	Graph-based \cite{wan14} & 0.565 \\
    \hline 
	FFT \cite{has16} & 0.771 \\
    \hline 
    \textbf{Proposed RLF} &  \textbf{0.783} \\
	\hline 
  \end{tabular}
  \caption{Performance evaluation of feature representation methods on BH2M Dataset.}
 \label{tab:feat2}
\end{table}

In order to highlight the importance of the proposed RLF descriptor, a comparison is done with the existing feature representation methods such as SIFT \cite{low04}, SURF \cite{bay08}, BRISK \cite{leu11}, Oriented FAST and Rotated BRIEF (ORB) \cite{rub11}, KAZE \cite{alc12}, DoLF \cite{zag17}, HoG \cite{alm12a}, Loci features \cite{fer11}, graph-based \cite{wan14} and FFT \cite{has16}. Table \ref{tab:feat1} presents the experimental results to evaluate the feature representation methods used in the word spotting framework for the Bentham dataset (ICFHR 2014 competition), as an example. This is with reference to the mAP values published in a recent work \cite{zag17} under the given experimental set up. It is observed from Table \ref{tab:feat1} that the RLF descriptor achieves higher mAP in comparison with SIFT, SURF, BRISK, ORB and KAZE, and performs comparable against DoLF. Table \ref{tab:feat2} validates the performance of the RLF descriptor with respect to the BH2M dataset, and the experiments are performed under the same test settings where the matching algorithm is same for all feature representation methods. The RLF descriptor performs significantly in comparison with other methods, because of the advantages inherited from relaxed feature representation and efficient algorithm design. Nevertheless, with reference to the three historical handwritten datasets used in the experiments, the proposed method is observed to be most consistent and stable with high mAP. %The Matlab implementation of this work can be found at \url{http://bit.ly/2IpR56o}.

%\begin{table}[htb]
%	\centering
%	\begin{tabular}{|l|l|l|l|}
%	\hline
%	Graphics & Top & In-between & Bottom \\
%	\hline
%	Tables & End & Last & First \\
%	\hline
%	Figures & Good & Similar & Very well \\
%	\hline
%	\end{tabular}
%	\caption{Table captions should be placed below the table}
%\end{table}

%{\bfseries Total length of a paper is max. 8 pages.}

%Please, use the standard Journal of WSCG format for references -- that is, a numbered list at the end of the article, ordered alphabetically by first author, and referenced by a name in brackets \cite{con00a}. See the examples of citations at the end of this document. Within this template file, use the style named references for the text of your citation.

%\begin{figure}[htb]
%    \centering
%    \rule{6cm}{3cm}
%    \caption{Insert caption to place caption below figure.}
%    \label{fig:box}
%\end{figure}

%\begin{table}[htb]
%	\centering
%	\begin{tabular}{|l|l|l|l|}
%	\hline
%	Graphics & Top & In-between & Bottom \\
%	\hline
%	Tables & End & Last & First \\
%	\hline
%	Figures & Good & Similar & Very well \\
%	\hline
%	\end{tabular}
%	\caption{Table captions should be placed below the table}
%\end{table}

%\section{Figures/Captions}
%Place Tables/Figures/Images in text as close to the reference as possible (see Figure\ref{fig:box}). It may extend across both columns to a maximum width of 16 cm (6.3"). Captions should be Times New Roman 10-points.  They should be numbered (e.g., "Table 1" or "Figure 2"), please note that the word for Table and Figure are spelled out. Figure's and Table's captions should be centered beneath the image, picture or a table.

\section{Conclusion}
\label{sec:con}
This paper presented a fast and robust Radial Line Fourier descriptor, with a short feature vector of 32 dimensions, for segmentation-free and training-free handwritten word spotting. A simple preconditioner-based feature matching algorithm is employed, and the experimental results on a variety of historical document images from well-known datasets demonstrate the effectiveness of the proposed method. Under the experimental settings, the proposed RLF descriptor based method outperformed the state-of-the-art methods, including the winners of the popular keyword spotting competitions. As future work, the ideas presented herein will be scaled to aid word feature representation for heavily degraded archival databases with improvements using query expansion.

%\section{Acknowledgments}
%This work was supported by the Swedish strategic research programme eSSENCE and the Riksbankens Jubileumsfond (Dnr NHS14-2068:1). The computations were performed on resources provided by SNIC through Uppsala Multidisciplinary Center for Advanced Computational Science (UPPMAX) under Project SNIC 2017/7-97. 

%-------------------------------------------------------------------------
% example of algorithm typesetting
% to allow this, uncomment line 
% \RequirePackage[noend]{myalgorithm}
% in the wscg.sty file
% and download that package from Gabriel Zachmann's page http://zach.in.tu-clausthal.de/latex/
%
%
%\begin{algorithm}
%\hrule
%  \centering
%\begin{algorithmic}
%    \STMT $d_{l,r} = f_B(P_1), f_B(P_n)$
%    \WHILE{ $|d_l| > \epsilon $ and $|d_r| > \epsilon $ and $l<r$}
%        \STMT $d_x = f_B(P_x)$
%        \IF{ $d_x < 0$ }
%            \STMT $l, r = x, r$
%        \ELSE
%            \STMT $l, r = l, x$
%        \ENDIF
%    \ENDWHILE
%\end{algorithmic}
%\hrule
%\caption{Example of some pseudo-code}
%\label{fg:code}
%\end{algorithm}

%-------------------------------------------------------------------------

%{\bfseries
%Last page should be fully used by text, figures etc. Do not leave empty space, please. 
%
%Do not lock the PDF -- additional text and info will be inserted, i.e. ISSN/ISBN etc. 
%}

\end{document}